\documentclass[letterpaper]{article} 
\usepackage{aaai23}  
\usepackage{times}  
\usepackage{helvet}  
\usepackage{courier}  
\usepackage[hyphens]{url}  
\usepackage{graphicx} 
\usepackage{natbib}  
\usepackage{caption} 
\usepackage{algorithm}
\usepackage{algorithmic}

\usepackage{newfloat}
\usepackage{listings}
\DeclareCaptionStyle{ruled}{labelfont=normalfont,labelsep=colon,strut=off} 
\floatstyle{ruled}
\newfloat{listing}{tb}{lst}{}
\floatname{listing}{Listing}
\title{Pre-Training With Scientific Text Improves Educational Question Generation (Student Abstract)}
\author{
Hamze Muse\equalcontrib, 
Sahan Bulathwela\equalcontrib 
and Emine Yilmaz
}

\affiliations{
Centre for Artificial Intelligence, University College London \\
    Gower Street, London WC1E 6BT, UK \\
    \{hamze.muse.20, m.bulathwela, emine.yilmaz\}@ucl.ac.uk
}

\begin{document}

\maketitle

\begin{abstract}
With the boom of digital educational materials and scalable e-learning systems, the potential for realising AI-assisted personalised learning has skyrocketed. In this landscape, the automatic generation of educational questions will play a key role, enabling scalable self-assessment when a global population is manoeuvring their personalised learning journeys. We develop \textit{EduQG}, a novel educational question generation model built by adapting a large language model. Our initial experiments demonstrate that \textit{EduQG} can produce superior educational questions by pre-training on scientific text. 
\end{abstract}

\section{Introduction}
While digital learning resources are created in abundance 
\cite{truelearn}, providing related questions 
to these resources 
facilitates self-testing, a critical element of self-regulated learning.
Also, question-answering enables an intelligent tutor to reliably verify learner skill mastery,
making scalable educational question generation essential for democratising education \cite{ai_ed_demo,zhang2021review}. 
While existing language models are being used for question generation,
their utility 
to education is only being explored very recently \cite{wang2022towards}. In particular, pre-training large language models with educational text to improve question generation is an unexplored area. 
This work validates if additional training with educational text can improve questions generated in the educational context. We develop an experiment to adapt a large language model to test this and propose \emph{EduQG}, a novel model for educational question generation. Our initial comparisons with a baseline question generation model indicate that this additional training can improve performance.  

\begin{table*}[t!] \centering \small
\begin{tabular}{c|ccccc|ccc}
\hline
                & \multicolumn{5}{c}{Predictive Performance}                          & \multicolumn{3}{c}{Linguistic Quality}       \\
Model           & BLEU-1 $\uparrow$     & BLEU-2 $\uparrow$     & BLEU-3 $\uparrow$     & BLEU-4   $\uparrow$   & F1-Score  $\uparrow$        & Perplexity $\downarrow$       & Diversity $\uparrow$         & Grammar Errors $\downarrow$        \\

\hline
Leaf (Baseline) & 27.07          & 20.22          & 17.17          & {16.46} & 30.90          & \textbf{30.82} & 0.735          & \textbf{0.102} \\
EduQG (Ours)        & \textbf{29.19} & \textbf{21.69} & \textbf{18.03} & \textbf{16.76} & \textbf{33.18} & 34.36          & \textbf{0.749} & 0.122  \\       
\hline
\end{tabular}
\caption{Comparison of predictive performance and linguistic quality between Leaf (baseline) and EduQG (our proposal). The superior performance is indicated in \textbf{bold} face.}
\label{results}
\end{table*}
\section{Related Work}

Prior work mainly utilises i) rule-based and ii) neural-based models for question generation (QG), while neural approaches have dominated the state of the art on QG in different applications including intelligent tutoring \cite{zhang2021review}. 
When it comes to leveraging QG for education, Leaf system \cite{vachev2022leaf} is one of the latest proposed methods.
Leaf is a cutting-edge question generation system that fine-tunes a large language model for question generation and multiple-choice distracter generation. Due to the recency and relevance of the Leaf system, we use the QG model of Leaf as the baseline model of this study.
Like many cutting-edge models, Leaf uses the \emph{SQuAD 1.1} \cite{DBLP:rajpurkar2016squad}, a reading comprehension dataset containing more than 100,000 questions crowd-sourced on a number of Wikipedia articles, to train the QG component of the system. It does so by fine-tuning a pre-trained T5 language model \cite{raffel2020exploring}. 
 
Although Leaf has been built for educational use cases using the SQuAD dataset, the SQuAD dataset itself contains questions that are aimed at English reading comprehension. Thus, it is not a strong candidate for testing the question generation capability for more rigorous subject domains such as the sciences. 
On the contrary, SciQ \cite{welbl-etal-2017-crowdsourcing} is a collection of 13,679 crowdsourced scientific exam questions that includes questions regarding physics, chemistry and other sciences. Although small in comparison to SQuAD, the SciQ dataset is a more relevant dataset that can be used to evaluate the educational QG capabilities of a model. Therefore, we use the SciQ dataset to evaluate the question generation models we build in this work. 

While large language models capture a lot of information about the world \cite{raffel2020exploring}, these models need to be pre-trained further in domain-specific datasets to improve their knowledge and fluency in specific domains (e.g. medicine \cite{https://doi.org/10.48550/arxiv.2109.04588}). In the realm of scientific information, \emph{S2ORC} is a corpus that consists of 81.1 million scholarly publications in English from various academic fields bringing together the largest publicly accessible collection of machine-readable academic literature to date \cite{lo-etal-2020-s2orc}. To test our hypothesis that pretraining the model with scientific/academic text would improve its educational QG capability, we use the S2ORC dataset.

\section{Our Approach}

\begin{figure}[]
    \centering
    \includegraphics[width=\columnwidth]{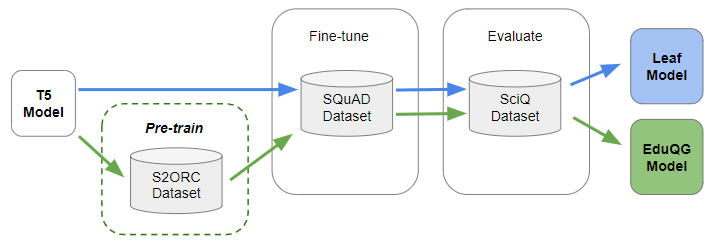}
    \caption{Baseline (blue arrows) and EduQG (green arrows).}
    \label{fig:method}
\end{figure}

The primary objective of our study is to validate if further fine-tuning a system on educational data can improve educational QG. The experiment we set up is illustrated in figure \ref{fig:method}. The foundational language model to both our training settings is the T5 language model \cite{raffel2020exploring}. We first replicate the QG component of the Leaf system \cite{vachev2022leaf} by taking the T5 model and fine-tuning it on the SQuAD 1.1 dataset as our baseline QG system (Blue flow in figure \ref{fig:method}). As the enhanced proposal, we use the same procedure, except, we fine-tune the T5 model with a down-sampled version of the S2ORC dataset  that contains approx. {23.2M} scientific abstracts related to Chemistry, Biology and Physics research papers (green dashed box in figure \ref{fig:method}). 

\paragraph{Evaluation} 
The two settings lead to the baseline (Leaf) and the proposed model (EduQG) that we compare using the SciQ dataset, as it contains exclusively educational questions.
To measure the predictive power of the human-generated questions, we use the BLUE score and the F1 score \cite{DBLP:rajpurkar2016squad}. To measure how human-like the generated questions are, we use perplexity, diversity and grammatical error rates. A lower perplexity score indicates better coherence \cite{wang2022towards}.

\section{Preliminary Results and Discussion}

The results of the model comparison are presented in table \ref{results}. 
The predictive performance results in Table \ref{results} clearly indicate that the \textit{EduQG} model is better at predicting scientific questions based on the context compared to Leaf. This is a strong indication that the additional scientific knowledge the EduGQ model is pre-trained on has an effect on educational QG capability. However, the linguistic quality metrics (shown on the right of the table) do not yield a favourable result although diversity has been improved by our model. We hypothesise that this may be due to the mismatch of language style and vocabulary of a scientific language that is advanced and complex. Therefore, scientific language might not align seamlessly with the reference models used for linguistic quality assessment. 

\section{Conclusion}
This work introduces EduQG, a foundational step toward further pre-training to improve educational QG. Our initial experiments prove the utility of pre-training an existing language model to improve its performance. The linguistic quality metrics are not as favourable as expected. Deeper analyses are warranted to understand whether the outcomes portray a limitation or a mismatch between the language models which will be addressed in future work using both offline and human studies. 

\subsection{Acknowledgments}
This work is funded by the European Commission project "Humane AI" (grant 820437).

\bibliography{aaai23}

\end{document}